\documentclass[10pt,twocolumn,letterpaper]{article}

\usepackage{cvpr}
\usepackage{times}
\usepackage{epsfig}
\usepackage{graphicx}
\usepackage{amsmath}
\usepackage{amssymb}
\usepackage{amsthm}
\usepackage{enumerate}
\usepackage{multirow}
\usepackage[FIGTOPCAP]{subfigure}
\usepackage{balance}
\usepackage{bbm}
\usepackage{color}
\usepackage{tabu}
\usepackage{makecell}
\usepackage{cite}

\usepackage[breaklinks=true,bookmarks=false]{hyperref}

\cvprfinalcopy 

\def\cvprPaperID{4107} 

\ifcvprfinal\fi
\begin{document}

\title{Leveraging the Invariant Side of Generative Zero-Shot Learning}

\author{Jingjing Li$^1$, Mengmeng Jing$^1$, Ke Lu$^1$, Zhengming Ding$^2$, Lei Zhu$^3$, Zi Huang$^4$\\
$^1$ University of Electronic Science and Technology of China; $^3$ Shandong Normal Unversity\\
$^2$ Indiana University-Purdue University Indianapolis; $^4$ University of Queensland\\
{\tt\small lijin117@yeah.net}
}

\maketitle
\thispagestyle{empty}

\begin{abstract}
   Conventional zero-shot learning (ZSL) methods generally learn an embedding, e.g., visual-semantic mapping, to handle the unseen visual samples via an indirect manner. In this paper, we take the advantage of generative adversarial networks (GANs) and propose a novel method, named leveraging invariant side GAN (LisGAN), which can directly generate the unseen features from random noises which are conditioned by the semantic descriptions. Specifically, we train a conditional Wasserstein GANs in which the generator synthesizes fake unseen features from noises and the discriminator distinguishes the fake from real via a minimax game. Considering that one semantic description can correspond to various synthesized visual samples, and the semantic description, figuratively, is the soul of the generated features, we introduce soul samples as the invariant side of generative zero-shot learning in this paper. A soul sample is the meta-representation of one class. It visualizes the most semantically-meaningful aspects of each sample in the same category. We regularize that each generated sample (the varying side of generative ZSL) should be close to at least one soul sample (the invariant side) which has the same class label with it. At the zero-shot recognition stage, we propose to use two classifiers, which are deployed in a cascade way, to achieve a coarse-to-fine result. Experiments on five popular benchmarks verify that our proposed approach can outperform state-of-the-art methods with significant improvements \footnote{Codes and datasets are available at github.com/lijin118/LisGAN}.
\end{abstract}

\section{Introduction}
In general, a computer vision algorithm can only handle the objects which appeared in the training dataset. In other words, an algorithm can only recognize the objects which are seen before. However, for some specific real-world applications, we either do not have the training sample of one object or the sample is too expensive to be labeled. For instance, we want the approach to trigger a message when it encounters a sample with a rare gene mutation from one species. Unfortunately, we did not have the visual features of the sample for training. The things we have are merely the images taken from normal instances and some semantic descriptions which describe the characteristics of the mutation and how it differs from normal ones. Conventional machine learning algorithms would fail in this task, but a human being would not. A human being is able to recognize an unseen object at the first glance by only reading some semantic descriptions. Inspired by this, zero-shot learning (ZSL)~\cite{ye2017zero,dinglow2017,verma2018generalized,xian2017zero} is proposed to handle unseen objects by the model which is trained on only seen objects and semantic descriptions about both seen and unseen categories. 

\begin{figure}[t]
\begin{center}
\includegraphics[width=0.85\linewidth]{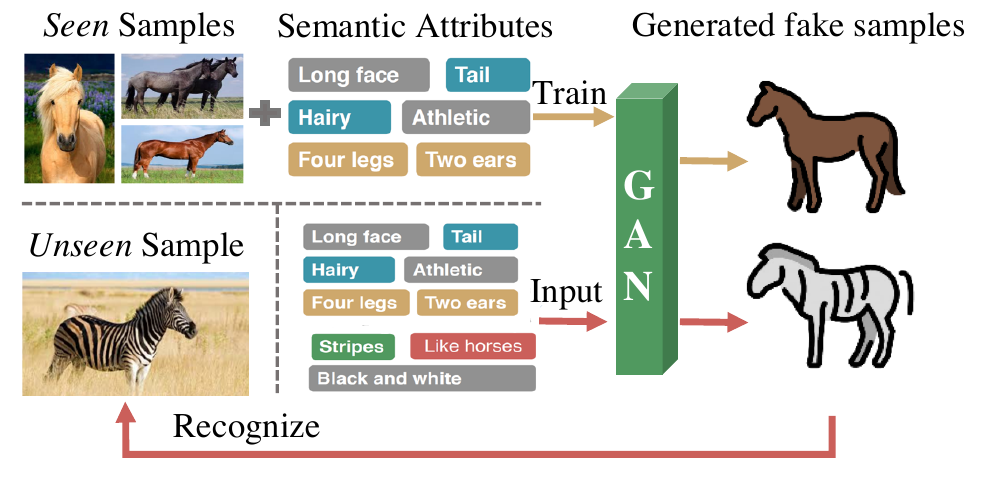}
\end{center}
\vspace{-10pt}
\caption{Zero-shot learning with GANs, i.e., generative ZSL.}
\label{fig:idea}
\vspace{-10pt} 
\end{figure}

Since the seen and unseen classes are connected by the semantic descriptions, a natural idea is to learn a visual-semantic mapping so that both seen and unseen samples can be compared in the semantic space. For instance, previous works~\cite{zhang2017learning,zhang2015zero,dinglow2017,zhang2016zero} learn either shallow or deep embeddings for zero-shot learning. These methods handle the unseen samples via an indirect way. Considering that one semantic description can correspond to enormous number of visual samples, the performance of zero-shot learning is restricted with the limited semantic information.

Recently, thanks to the advances in generative adversarial networks (GANs)~\cite{goodfellow2014generative}, a few approaches are proposed to directly generate unseen samples from the random noises and semantic descriptions~\cite{xian2018feature,zhu2018generative,mishra2017generative}, as shown in Fig.~\ref{fig:idea}. With the generated unseen samples, zero-shot learning can be transformed to a general supervised machine learning problem. In such a learning paradigm, however, the challenges of zero-shot learning have been also passed on to the GANs. In the GANs based paradigms for zero-shot learning, we have to address the spurious and soulless generating problem. Specifically, we generally have only one semantic description, e.g., one attributes vector, one article or one paragraph of texts, for a specific category, but the semantic description is inherently related to a great mass of images in the visual space. For instance, ``a tetrapod with a tail'' can be mapped to many animals, e.g., cats, dogs and horses. At the same time, some objects from different categories have very similar attributes, such as ``tigers'' and ``ligers''. Thus, the generative adversarial networks for zero-shot learning must challenge two issues: 1) {\it how to guarantee the generative diversity based on limited and even similar attributes?} 2) {\it how to make sure that each generated sample is highly related with the real samples and corresponding semantic descriptions?} However, since deploying GANs to address the ZSL problem is a new topic, most of existing works did not explicitly address the two issues. In this paper, we propose a novel approach which takes the two aspects into consideration and carefully handles them in the formulation. 

At first, to guarantee that the generated samples are meaningful, we propose to generate samples from random noises which are conditioned with the class semantic descriptions. At the same time, we also introduce the supervised classification loss in the GAN discriminator to preserve the inter-class discrimination during the adversarial training. Furthermore, to ensure that each synthesized sample (the varying side of generative zero-shot learning) is highly related with the real ones and corresponding semantic descriptions (the invariant side), we introduce soul samples in this paper, as shown in Fig.~\ref{fig:soul}. For unseen classes, the visual characteristics of a generated sample only depend on the semantic descriptions. Thus, the semantic information is the soul of the generated samples. The soul sample must be not very specific so that it can plainly visualize the most semantically-meaningful aspects and relate to as many samples as possible. For the seen images, therefore, we define that a soul sample is an average representation of them. For the generated samples, we regularize them to be close to soul samples. Thus, we can guarantee that each generated sample is highly related with the real ones and corresponding semantic descriptions. 

\begin{figure*}[t]
\begin{center}
\includegraphics[width=0.96\linewidth]{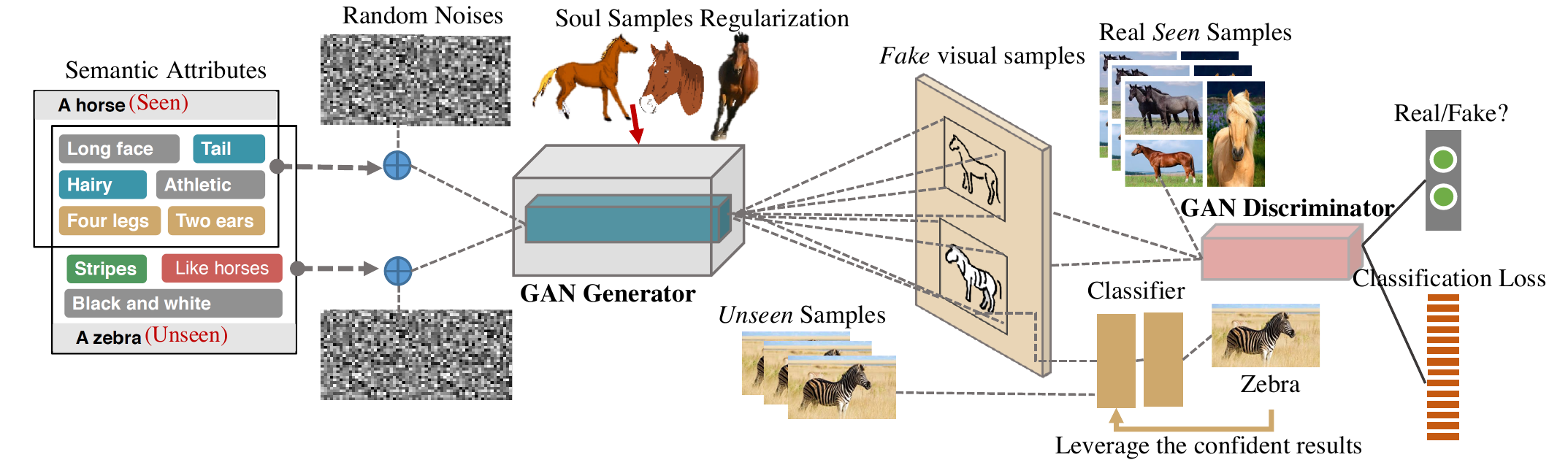}
\end{center}
\vspace{-10pt}
\caption{Idea illustration of our LisGAN (Leveraging invariant side GAN). We train a conditional WGAN to generate fake unseen images from random noises and semantic attributes. Multiple soul samples for each class are introduced to regularize the generator. Unseen samples classified with high confidence are leveraged to fine-tune final results.}
\label{fig:ideaill}
\vspace{-10pt} 
\end{figure*}

To summarize, the main contributions of this paper are:
\begin{enumerate}[1)]
 \item We propose a novel ZSL method LisGAN which takes advantage of generative adversarial networks. Specifically, we deploy the conditional GANs to tackle the two issues: generative diversity and generative reliability. To improve the quality of generated features, we introduce soul samples which are defined as the representations of each category. By further considering the multi-view nature of different images, we propose to define multiple soul samples for each class. We regularize each generated sample to be close to at least one soul sample so that the varying side in generative zero-shot learning would not be divorced from the invariant side.

\item At the zero-shot recognition stage, we propose that if we have high confidence in recognizing an unseen sample, the sample (with its assigned pseudo label) will be leveraged as the reference to recognize other unseen samples. Specifically, we propose to use two classifiers, which are deployed in a cascade way, to achieve a coarse-to-fine result. We also report a simple yet efficient method to measure the classification confidence in this paper. 

\item Extensive experiments on five widely used datasets verify that our proposed method can outperform state-of-the-art methods with remarkable improvements.
\end{enumerate}


\section{Related Work}
\subsection{Zero-Shot Learning}

Inspired by the human ability that one can recognize an object at the first glance by only knowing some semantic descriptions of it, zero-shot learning~\cite{xian2017zero,changpinyo2016synthesized,kodirov2017semantic,li2019zero,ding2018generative} aims to learn a model with good generalization ability which can recognize unseen objects by only giving some semantic attributes. A typical zero-shot learning model is trained on visual features which only contain the seen samples and semantic features which contain both seen and unseen samples. Since the seen objects and unseen ones are only connected in the semantic space and the unseen objects need to be recognized by the visual features, zero-shot learning methods generally learn a visual-semantic embedding with the seen samples. At the zero-shot classification stage, unseen samples are projected into the semantic space and labeled by semantic attributes~\cite{lei2015predicting,lampert2014attribute,romera2015embarrassingly,dinglow2017}. Instead of learning a visual-semantic embedding, some previous works also propose to learn a semantic-visual mapping so that the unseen samples can be represented by the seen ones~\cite{kodirov2015unsupervised,shigeto2015ridge}. In addition, there are also some works to learn an intermediate space shared by the visual features and semantic features~\cite{changpinyo2016synthesized,zhang2015zero,zhang2016zero}. Besides, ZSL is also related with domain adaptation and cold-start recommendation~\cite{li2018transfer,li2018heterogeneous,li2018I,li2017two}. 

From the recent literatures, typical zero-shot learning tasks are zero-shot classification~\cite{jiang2017learning,ye2017zero}, zero-shot retrieval~\cite{long2018towards} and generalized zero-shot recognition~\cite{verma2018generalized}. The main difference between zero-shot learning and generalized zero-shot recognition is that the former only classifies the unseen samples in the unseen category and the latter recognizes samples, which can be either seen ones and unseen ones, in both seen and unseen categories.

It is easy to observe that conventional zero-shot learning methods are {\it indirect}. They usually need to learn a space mapping function. Recently, by taking advantage of generative adversarial networks~\cite{arjovsky2017wasserstein,goodfellow2014generative}, several methods~\cite{zhu2018generative,xian2018feature} are proposed to {\it directly} generate unseen samples from their corresponding attributes, which converts the conventional zero-shot learning to a classic supervised learning problem.

\subsection{Generative Adversarial Nets}
A typical generative adversarial networks (GANs)~\cite{goodfellow2014generative} consists of two components: a generator and a discriminator. The two players are trained in an adversarial manner. Specifically, the generator $G$ tries to generate fake images from input noises to fool the discriminator, while the discriminator $D$ attempts to distinguish real images and fake ones. In general, the input of $G$ is random noise and the output is the synthesized image. The inputs of $D$ are both real images and fake images, the output is a probability distribution. In this paper, we deploy $G$ to generate sample features instead of image pixels.

Although GANs has shown quite impressive results and profound impacts, the vanilla GAN is very hard to train. Wasserstein GANs (WGANs)~\cite{arjovsky2017wasserstein} presents an alternative to traditional GAN training. WGANs can improve the stability of learning, get rid of problems like mode collapse, and provide meaningful learning curves useful for debugging and hyperparameter searches. In addition, conditional GANs~\cite{mirza2014conditional} are proposed to enhance the outputs of traditional GANs. With conditional GANs, one can incorporate the class labels and other information into the generator and discriminator to synthesize specified samples.

\section{The Proposed Method}
\subsection{Definitions and Notations}
Given $n$ labeled seen samples with both visual features $X\in \mathbb{R}^{d\times n}$ and semantic descriptions $A\in \mathbb{R}^{m\times n}$ for training, zero-shot learning aims to recognize $n_u$ unknown visual samples $X_u\in \mathbb{R}^{d\times n_u}$ which only have semantic attributes $A_u\in \mathbb{R}^{m\times n_u} $ for training. Let $Y$ and $Y_u$ be the label space of $X$ and $X_u$, respectively, in zero-shot learning we have $Y \cap Y_u=\varnothing$. Suppose that we have $C$ and $C_u$ categories in total for seen data and unseen data, respectively, classical zero-shot learning recognizes $X_u$ by only searching in $C_u$, while generalized zero-shot learning searches in $C \cup C_s$. The semantic descriptions $A$ and $A_u$ are either provided as binary/numerical vectors or word embedding/RNN features. Each semantic description $a$ corresponds to a category $y$.
Formally, given \{$X$, $A$, $Y$\} and \{$A_u$, $Y_u$\} for training, the task of zero shot learning is to learn a function $f: \mathcal{X}_u \rightarrow \mathcal{Y}_u$ and generalized zero shot learning is to learn a function $f: \{\mathcal{X, X}_u\} \rightarrow \mathcal{Y} \cup \mathcal{Y}_u$ .

\subsection{Overall Idea}
In this paper, we take advantage of GANs to directly generate fake visual features for unseen samples from random noises and the semantic descriptions. Then, the synthesized visual features are used as references to classify real unseen samples. Since we only have $A_u$ and the GAN discriminator cannot access $X_u$ in the training stage, the real or fake game, therefore, cannot be played. Thus, we mainly train our GAN on the seen classes. At the same time, we deploy the conditional GANs so that the class embedding can be incorporated into both generator $G$ and discriminator $D$. Since \{$A$, $Y$\} and \{$A_u$, $Y_u$\} are interconnected, i.e., $A$ and $A_u$ have the same semantic space, the conditional GAN which generates high-quality samples for seen classes is also expected to generate high-quality samples for unseen categories. The main idea of this paper is illustrated in Fig~\ref{fig:ideaill}. Compared with existing methods which also deploy GANs for zero-shot learning, our novelty comes from two aspects. The first one is that we introduce multiple soul samples per class to regularize the generator. The second is that we leverage the unseen samples which are classified with high confidence to facilitate the subsequent unseen samples. Experiments reported in section~5 show that we can achieve a significant improvement against state-of-the-art methods on various datasets.

\subsection{Train the LisGAN}

Given the seen samples \{$X$, $A$, $Y$\}, the attributes $A_u$ of the unseen sample and random noises $z\sim \mathcal{N}(0,1)$, the GAN generator $G$ uses the input $a$ and nosies $z$ to synthesize fake features. At the same time, the GAN discriminator $D$ takes the features of real image $x$ and $G(z,a)$ as inputs to discriminate whether an input feature is real or fake. Formally, the loss of $G$ can be formulated as follows:
\begin{equation}
\label{eq:gen}
  \begin{array}{c}
 L_G= -\mathbb{E}[D(G(z,a))]- \lambda \mathbb{E}[\mathrm{log}P(y|G(z,a))],
  \end{array} 
\end{equation} 
where the first term is the Wasserstein loss~\cite{arjovsky2017wasserstein} and the second one is the supervised classification loss on the synthesized features, $\lambda>0$ is a balancing parameter. 

Similarly, the loss of the discriminator can be formulated as follows:
\begin{equation}
\label{eq:dis}
  \begin{array}{l}
 L_D= \mathbb{E}[ D(G(z,a))]-\mathbb{E}[ D(x)] \\
 ~~~~~~~~~~- \lambda (\mathbb{E}[\mathrm{log} P(y|G(z,a))] + \mathbb{E}[\mathrm{log} P(y|x)]) \\
 ~~~~~~~~~~- \beta \mathbb{E}[(\|\nabla_{\hat{x}} D(\hat{x})\|_2-1)^2],
  \end{array} 
\end{equation} 
where $\beta>0$ is a hyper-parameter. The fourth term, similar with the third one, is a supervised classification loss on real samples. The last term is used to enforce the Lipschitz
constraint~\cite{gulrajani2017improved}, in which $\hat{x}=\mu x +(1-\mu)G(z,a)$ with $\mu \sim U(0,1)$. As suggested in~\cite{gulrajani2017improved}, we fix $\beta=10$.

\begin{figure}[t]
\begin{center}
\vspace{-10pt} 
\includegraphics[width=0.85\linewidth]{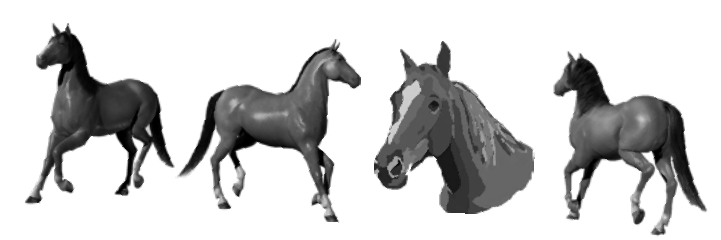}
\end{center}
\vspace{-10pt}
\caption{Soul samples of the horse category. Considering the nature multi-view property of visual objects, e.g., real images of an object are usually captured from different views, we propose to learn multiple soul samples for each class. By such a formulation, the domain shift issue caused by different views can be alleviated.}
\label{fig:soul}
\vspace{-5pt} 
\end{figure}

In our model, we take the CNN features of samples as the visual input $X$. Both the generator and discriminator are implemented with fully connected layers and ReLU activations. Thus, the model is feasible to incorporate into different CNN architectures. At the same time, the output of the generator is directly visual features rather than image pixels. By optimizing the above two-player minimax game, the conditional GAN generator is able to synthesize fake features of the seen images with the class embedding $A$. Since the unseen objects share the same semantic space with the seen samples, the conditional GAN generator can also synthesize visual features for unseen categories via $A_u$.  With the optimization problem in Eq.~\eqref{eq:gen} and Eq.~\eqref{eq:dis}, our model can guarantee the generative diversity with similar attributes. With the supervised classification loss, it can also ensure that the learned features are discriminative for further classification. However, the model does not explicitly address the quality of the generated features. In this paper, to make sure that each generated feature is highly related with the semantic descriptions and real samples, we introduce soul samples to regularize the generator. Since the soul samples of a category should reflect the most remarkable characteristics of the class as much as possible, we deploy the average representation of all samples from the category $c$ to define the soul sample of $c$, which is similar with prototypical networks for few-shot learning~\cite{snell2017prototypical}. Furthermore, considering the nature multi-view property of real samples, as shown in Fig.~\ref{fig:soul}, we further propose that a category $c$ should have multiple soul samples to address the multi-view issue. To this end, we first group the real features of one seen class into $k$ clusters. For simplicity, we fix $k=3$ in this paper. Then, we calculate a soul sample for each cluster. Let $\{X_1^c, X_2^c, \cdots, X_k^c\}$ be the $k$ clusters of category $c$, the soul samples $S^c=\{s_1^c, s_2^c, \cdots, s_k^c\}$ are defined as:
\begin{equation}
\label{eq:soul1}
  \begin{array}{c}
 s_k^c= \frac{1}{|X_k^c|} \sum\limits_{x_i\in X_k^c} x_i.
  \end{array} 
\end{equation} 

Similarly, for the generated fake features, we can also define the soul sample $\tilde{s}_k^c$ as:

\begin{equation}
\label{eq:soul2}
  \begin{array}{c}
 \tilde{s}_k^c= \frac{1}{|\tilde{X}_k^c|} \sum\limits_{\tilde{x}_i\in \tilde{X}_k^c} \tilde{x}_i ,
  \end{array} 
\end{equation} 
where $\tilde{x}_i=G(z,a)$ is a generated fake feature.

In this paper, we encourage that each generated sample $\tilde{x}$ for class $c$ should be close to at least one soul sample $s^c$. Formally, we introduce the following regularization:  
\begin{equation}
\label{eq:regu1}
  \begin{array}{c}
 L_{R1}= \frac{1}{n_1}\sum\limits_{i=1}^{n_1} \min\limits_{j\in[1,k]} \|\tilde{x}_i - s_j^c\|_2^2,
  \end{array} 
\end{equation} 
where $n_1$ is the number of generated samples and $k$ is the number of soul samples per class. At the same time, since the soul samples can also be seen as the centroid of one cluster, we encourage that the fake soul samples should be close to at least one real soul sample from the same class, which can be formulated as:
\begin{equation}
\label{eq:regu2}
  \begin{array}{c}
 L_{R2}= \frac{1}{C}\sum\limits_{c=1}^{C} \min\limits_{j\in[1,k]} \|\tilde{s}_j^c - s_j^c \|_2^2,
  \end{array} 
\end{equation} 
where $C$ is the number of total categories. With the two regularizations $L_{R1}$ and $L_{R2}$, our model avoids to generate soulless features. Each of the generated features would be close to the real ones, which guarantees the quality of the fake features. From another perspective, $L_{R1}$ is an individual regularization which addresses single samples and $L_{R2}$ is a group regularization which takes care of a cluster. 

\subsection{Predict Unseen Samples}
Once the GAN is trained to be able to generate visual features for seen classes, it can also synthesize visual features for the unseen ones with random noises and semantic attributes $A_u$. Then, the zero-shot learning is automatically converted to a supervised learning problem. Specifically, we can train a softmax classifier on the generated features and classify the real unseen features. The softmax is formulated as minimizing the following negative log likelihood:
\begin{equation}
\label{eq:softmax1}
  \begin{array}{c}
 \min\limits_\theta -\frac{1}{|\mathcal{X}|} \sum\limits_{{(x,y)}\in \mathcal{(X,Y)}} \mathrm{log}P(y|x;\theta),
  \end{array} 
\end{equation} 
where $\theta$ is the training parameter and 
\begin{equation}
\label{eq:softmax2}
  \begin{array}{c}
 P(y|x;\theta)= \frac{\mathrm{exp}(\theta_y^\top x)}{\sum_{i=1}^{N}\mathrm{exp}(\theta_i^\top x)}.
  \end{array} 
\end{equation} 

In this paper, we further propose that we can leverage an unseen sample if we have sufficient confidence in believing that the sample has been correctly classified. Since the output of the softmax layer is a vector which contains the probabilities of all possible categories, the entropy of the vector can be used to measure the certainty of the results. If a probability vector has lower entropy, we have more confidence of the results. Therefore, we leverage the samples which have low classification entropy and deploy them as references to classify the other unseen samples. Specifically, we calculate the sample entropy by:
\begin{equation}
\label{eq:sample-entropy}
  \begin{array}{c}
 \mathrm{E}({y})= -\sum\limits_{c=1}^{C} y_c~\mathrm{log}~y_c.
  \end{array} 
\end{equation}

In our model, we deploy two classifiers via a cascade manner to predict the unseen samples. The first classifier is used to evaluate the classification confidence and the second is used to leverage the correctly classified samples. In our zero-shot recognition, the first classifier is a softmax trained on the generated fake features, while second classifier can be either a trained classifier, e.g., softmax classifier, SVM, or just a training-free classifier, e.g., NNC. 





\section{Experiments}

\subsection{Datasets}
APascal-aYahoo ({\bf aPaY}) contains 32 categories from both PASCAL VOC 2008 dataset and Yahoo image search engine. Specifically, 20~classes are from PASCAL and 12~classes are from Yahoo. The total number of aPaY is 15,339. Following previous work~\cite{xian2017zero,zhu2018generative}, we deploy the PASCAL VOC 2008 as seen dataset and the Yahoo as unseen one. An additional 64-dimensional attribute vector is annotated for each category.

Animals with Attributes ({\bf AwA})~\cite{lampert2009learning} consists of 30,475 images of $50$ animals classes. The animals classes are aligned with Osherson's classical class/attribute matrix, thereby providing 85 numeric attribute values for each class.

Caltech-UCSD Birds-200-2011 ({\bf CUB})~\cite{wah2011caltech} is an extended version of the CUB-200 dataset. CUB is a challenging dataset which contains 11,788 images of 200 bird species. Each species is associated with a Wikipedia article and organized by scientific classification (order, family, genus, species). A vocabulary of 28 attribute groupings and 312 binary attributes were associated with the dataset based on an online tool for bird species identification. 

Oxford Flowers ({\bf FLO})~\cite{Nilsback08} dataset consists of 8,189 images which comes from 102 flower categories. Each class consists of between 40 and 258 images. The images have large scale, pose and light variations. In addition, there are categories that have large variations within the category and several very similar categories. For this dataset, we use the same semantic descriptions provided by Reed et al.~\cite{reed2016learning}.

SUN attributes ({\bf SUN})~\cite{patterson2012sun} is a large-scale scene attribute dataset, which spans 717 categories and 14,340 images in total. Each category includes 102 attribute labels. 

For clarity, we report the dataset statistics and zero-shot split settings in Table~\ref{tab:dataset}. The zero-shot splits of aPaY, AwA, CUB and SUN are same with previous work~\cite{xian2017zero} and the splits of FLO is same with~\cite{reed2016learning}. For the real CNN features, we follow previous work~\cite{xian2018feature} to extract 2048-dimensional features from ResNet-101~\cite{he2016deep} which is pre-trained on ImageNet. For the semantic descriptions, we use the default attributes included in the datasets. Specifically, since FLO did not provide attributes with the dataset, we use the 1024-dimensional RNN descriptions via the model of~\cite{reed2016learning}. For fair comparisons, all of our experimental settings are same with the protocols reported in previous work~\cite{xian2018feature}.

\begin{table}[t!p]
\centering
\caption{Dataset statistics. The (number) in \# Seen Classes indicates the number of seen classes used for test in the GZSL. }
\label{tab:dataset}
\footnotesize\begin{tabular}{lccccc}
\Xhline{0.75pt}
Dataset & aPaY & AwA & CUB & FLO & SUN  \\
\hline
\# Samples & 15,339 & 30,475 & 11,788 & 8,189 & 14,340   \\
\# Attributes & 64 & 85 & 312 & 1,024 & 102  \\
\# Seen Classes & 20 (5) & \!\!40 (13)\!\! & \!\!150 (50)\!\! & 82 (20)\!\! & 645 (65)  \\     
\# Unseen Classes\!\!\!\!\! & 12 & 10 & 50 & 20 & 72  \\
\Xhline{0.75pt}
\end{tabular}
\vspace{-10pt}
\end{table}

\subsection{Implementation and Compared Methods}
In our model, the GAN is implemented via multilayer perceptron with Rectified Linear Unit (ReLU) activation. Specifically, the generator $G$ contains a fully connected layer with 4,096 hidden units. The noise $z$ is conditioned by the semantic description $a$ and then severed as the inputs of $G$. An additional ReLU layer is deployed as the output layer of $G$ which outputs the synthesized fake features. The discriminator $D$ takes the real features and the synthesized fake features from $G$ and processes them via an FC layer, a Leaky ReLU layer, an FC layer and a ReLU layer. The discriminator has two branches for output. One is used to tell fake from real and the other is a standard $n$-ways classifier to predict the correct category of each sample. In this paper, we set $\lambda=0.01$ and $\beta=10$. The weight for two regularizations are all set to $0.01$. The sample entropy threshold is set to be smaller than the median of all entropies. One can also tune the hyper-parameters by cross-validation.

The compared methods are representative ones published in the fast few years and the state-of-the-art ones reported very recently. Specifically, we compare our approach with: DAP~\cite{lampert2014attribute}, CONSE~\cite{norouzi2013zero}, SSE~\cite{zhang2015zero}, DeViSE~\cite{frome2013devise}, SJE~\cite{akata2015evaluation}, ESZSL~\cite{romera2015embarrassingly}, ALE~\cite{akata2016label}, SYNC~\cite{changpinyo2016synthesized}, SAE~\cite{kodirov2017semantic}, DEM~\cite{zhang2017learning}, GAZSL~\cite{zhu2018generative} and f-CLSWGAN~\cite{xian2018feature}.

Following previous work~\cite{xian2018feature,zhu2018generative}, we report the average per-class top-1 accuracy for each of the evaluated method. Specifically, for classic zero-shot learning, we report the top-1 accuracy of unseen samples by only searching the unseen label space. However, for the generalized zero-shot learning, we report the accuracy on both seen classes and unseen classes with the same settings in~\cite{xian2017zero}. Some of the results reported in this paper are also cited from~\cite{xian2017zero}.


\begin{table}[t!p]
\centering
\caption{The top-1 accuracy (\%) of zero-shot learning on different datasets. The best results are highlighted with bold numbers.} 

\label{tab:zsl}
\footnotesize\begin{tabular}{lccccc}
\Xhline{0.75pt}
Methods & ~~aPaY~~ & ~~AwA~~ & ~~CUB~~ & FLO & SUN  \\
\hline
DAP~\cite{lampert2014attribute} & 33.8 & 44.1 & 40.0 & - & 39.9   \\
CONSE~\cite{norouzi2013zero} & 26.9 & 45.6 & 34.3 & - & 38.8  \\
SSE~\cite{zhang2015zero} & 34.0 & 60.1 & 43.9 & - & 51.5  \\     
DeViSE~\cite{frome2013devise} & 39.8 & 54.2 & 52.0 & 45.9 & 56.5  \\

SJE~\cite{akata2015evaluation} & 32.9 & 65.6 & 53.9 & 53.4 & 53.7   \\
ESZSL~\cite{romera2015embarrassingly} & 38.3 & 58.2 & 53.9 & 51.0 & 54.5  \\
ALE~\cite{akata2016label} & 39.7 & 59.9 & 54.9 & 48.5 & 58.1  \\     
SYNC~\cite{changpinyo2016synthesized} & 23.9 & 54.0 & 55.6 & - & 56.3  \\

SAE~\cite{kodirov2017semantic} & 8.3 & 53.0 & 33.3 & - & 40.3   \\
DEM~\cite{zhang2017learning} & 35.0 & 68.4 & 51.7 & - & {\bf 61.9}  \\
GAZSL~\cite{zhu2018generative} & 41.1 & 68.2 & 55.8 & 60.5 & 61.3  \\     
f-CLSWGAN~\cite{xian2018feature} & 40.5 & 68.2 & 57.3 & 67.2 & 60.8  \\
\hline
LisGAN [Ours]  & {\bf 43.1} & {\bf 70.6} & {\bf 58.8} & {\bf 69.6} &  61.7  \\
\Xhline{0.75pt}
\end{tabular}
\end{table}

\begin{table}[t]
\centering
\caption{The results (top-1 accuracy \%) of generalized zero-shot learning on aPaY dataset. The Mean in this table is the harmonic mean of seen and unseen samples, i.e., Mean=(2*Unseen*Seen)/(Unseen+Seen).} 
\label{tab:gzsl1}
\footnotesize\begin{tabular}{lccc}
\Xhline{0.75pt}
\multirow{2}{*}{Methods} & \multicolumn{3}{c}{aPaY}   \\
\cline{2-4}
 & ~~~~Unseen~~~~ & ~~~~~Seen~~~~~ & ~~~~~Mean~~~~~\\
 \hline
DAP~\cite{lampert2014attribute} & 4.8 & 78.3 & 9.0   \\
CONSE~\cite{norouzi2013zero} & 0.0 & {\bf 91.2} & 0.0  \\
SSE~\cite{zhang2015zero} & 0.2 & 78.9 & 0.4  \\     
DeViSE~\cite{frome2013devise} & 4.9 & 76.9 & 9.2  \\

SJE~\cite{akata2015evaluation} & 3.7 & 55.7 & 6.9    \\
ESZSL~\cite{romera2015embarrassingly} & 2.4 & 70.1 & 4.6  \\
ALE~\cite{akata2016label} & 4.6 & 73.7 & 8.7  \\     
SYNC~\cite{changpinyo2016synthesized} & 7.4 & 66.3 & 13.3  \\

SAE~\cite{kodirov2017semantic} & 0.4 & {80.9} & 0.9   \\
DEM~\cite{zhang2017learning} & 11.1 & 75.1 & 19.4  \\
GAZSL~\cite{zhu2018generative} & 14.2 & 78.6 & 24.0  \\     
f-CLSWGAN~\cite{xian2018feature}~~~~~~~ & 32.9& 61.7 & 42.9  \\
\hline
LisGAN [Ours]  & {\bf 34.3} & 68.2 & {\bf 45.7}  \\
\Xhline{0.75pt}
\end{tabular}
\vspace{-10pt}
\end{table}

\subsection{Zero-shot Learning}

We report the zero-shot learning results on the five datasets in Table~\ref{tab:zsl}. In these experiments, the possible categories of unseen samples are searched from only $Y_u$. It can be seen that our method achieves the best on four of the five evaluations. We also achieved state-of-the-art result on the last dataset. Specifically, we achieved 2.6\% improvement over the state-of-the-art method on aPaY dataset. We also achieved 2.4\%, 1.5\% and 2.4\% on AWA, CUB and FLO.


From the results, we can also observe that the GAN-based methods, e.g., GAZSL, f-CLSWGAN and ours, generally perform better than embedding ones, e.g., SSE, ALE and SAE. The embedding methods handle the unseen samples via an indirect manner, while the GAN method directly handle it by converting it to a supervised learning task. The results suggest that GAN could be a promising way to address zero-shot learning problem in the future. Apart from generating visual features from noises, GANs can also be used for semantic augmentation in zero-shot learning. In our future work, we will incorporate semantic data augmentation in our model to cover more unseen samples.

\subsection{Generalized Zero-shot Learning}

We further report the experiment results of generalized zero-shot learning in Table~\ref{tab:gzsl1} and Table~\ref{tab:gzsl2}. Table~\ref{tab:gzsl1} shows the results on aPaY dataset and Table~\ref{tab:gzsl2} shows the results on the other $4$~datasets. In generalized zero-shot learning, the seen classes are split into two parts: one for training and the other for test. At the test stage, both seen and unseen samples are recognized by searching the possible categories from $Y\cup Y_u$. The splits of seen classes can be seen in Table~\ref{tab:dataset} and more details can be found in previous work~\cite{xian2017zero}. Since both seen and unseen classes are tested in generalized zero-shot learning, we also report the harmonic mean of seen accuracy and unseen accuracy in the tables.

\begin{table*}[t]
\centering
\caption{The results (top-1 accuracy \%) of generalized zero-shot learning. The Mean in this table is the harmonic mean of seen and unseen samples, i.e., Mean=(2*Unseen*Seen)/(Unseen+Seen). The best results are highlighted with bold numbers.} 
\label{tab:gzsl2}
\small\begin{tabular}{l|ccc|ccc|ccc|ccc}
\Xhline{0.75pt}
\multirow{2}{*}{Methods} & \multicolumn{3}{c|}{AwA} & \multicolumn{3}{c|}{CUB} & \multicolumn{3}{c|}{FLO} & \multicolumn{3}{c}{SUN}  \\
\cline{2-13}
 & Unseen & Seen & Mean & Unseen & Seen & Mean & Unseen & Seen & Mean & Unseen & Seen & Mean \\
 \hline 
DAP~\cite{lampert2014attribute} & 0.0 & 88.7 & 0.0 & 1.7 & 67.9 & 3.3 & - & - & - & 4.2 & 25.2 & 7.2  \\
CONSE~\cite{norouzi2013zero}  & 0.4 & 88.6 & 0.8 & 1.6 & 72.2 & 3.1 & - & - & - & 6.8 & 39.9 & 11.6\\
SSE~\cite{zhang2015zero} & 7.0 & 80.5 & 12.9 & 8.5 & 46.9 & 14.4 & - & - & - & 2.1 & 36.4 & 4.0  \\     
DeViSE~\cite{frome2013devise} & 13.4 & 68.7 & 22.4 & 23.8 & 53.0 & 32.8 & 9.9 & 44.2 & 16.2 & 16.9 & 27.4 & 20.9  \\

SJE~\cite{akata2015evaluation} & 11.3 & 74.6 & 19.6 & 23.5 & 59.2 & 33.6 & 13.9 & 47.6 & 21.5 & 14.7 & 30.5 & 19.8   \\
ESZSL~\cite{romera2015embarrassingly} & 5.9 & 77.8 & 11.0 & 2.4 & 70.1 & 4.6 & 11.4 & 56.8 & 19.0 & 11.0 & 27.9 & 15.8  \\
ALE~\cite{akata2016label} & 14.0 & 81.8 & 23.9 & 4.6 & 73.7 & 8.7 & 13.3 & 61.6 & 21.9 &21.8 & 33.1 & 26.3  \\     
SYNC~\cite{changpinyo2016synthesized} & 10.0 & {\bf 90.5} & 18.0 & 7.4 & 66.3 & 13.3 & - & - & - & 7.9 & {\bf 43.3} & 13.4  \\

SAE~\cite{kodirov2017semantic} & 1.1 & 82.2 & 2.2 & 0.4 & {\bf 80.9} & 0.9 & - & - & - & 8.8 & 18.0 & 11.8   \\
DEM~\cite{zhang2017learning} & 30.5 & 86.4 & 45.1 & 11.1 & 75.1 & 19.4 & - & - & - & 20.5 & 34.3 & 25.6  \\
GAZSL~\cite{zhu2018generative} & 19.2 & 86.5 & 31.4 & 23.9 & 60.6 & 34.3 & 28.1 & 77.4 & 41.2 & 21.7 & 34.5 & 26.7  \\     
f-CLSWGAN~\cite{xian2018feature} & {\bf 57.9} & 61.4 & 59.6 & 43.7 & 57.7 & 49.7 & {\bf 59.0} & 73.8 & 65.6 & {42.6} & 36.6 & 39.4  \\
\hline
LisGAN [Ours]  & 52.6 & 76.3 & {\bf 62.3} & {\bf 46.5} & 57.9 & {\bf 51.6} & {57.7} & {\bf 83.8} & {\bf 68.3} & {\bf 42.9} & 37.8 & {\bf 40.2}  \\
\Xhline{0.75pt}
\end{tabular}
\end{table*}

\begin{figure}[t]
\begin{center}
\subfigure[f-CLSWGAN]{
\includegraphics[width=0.51\linewidth]{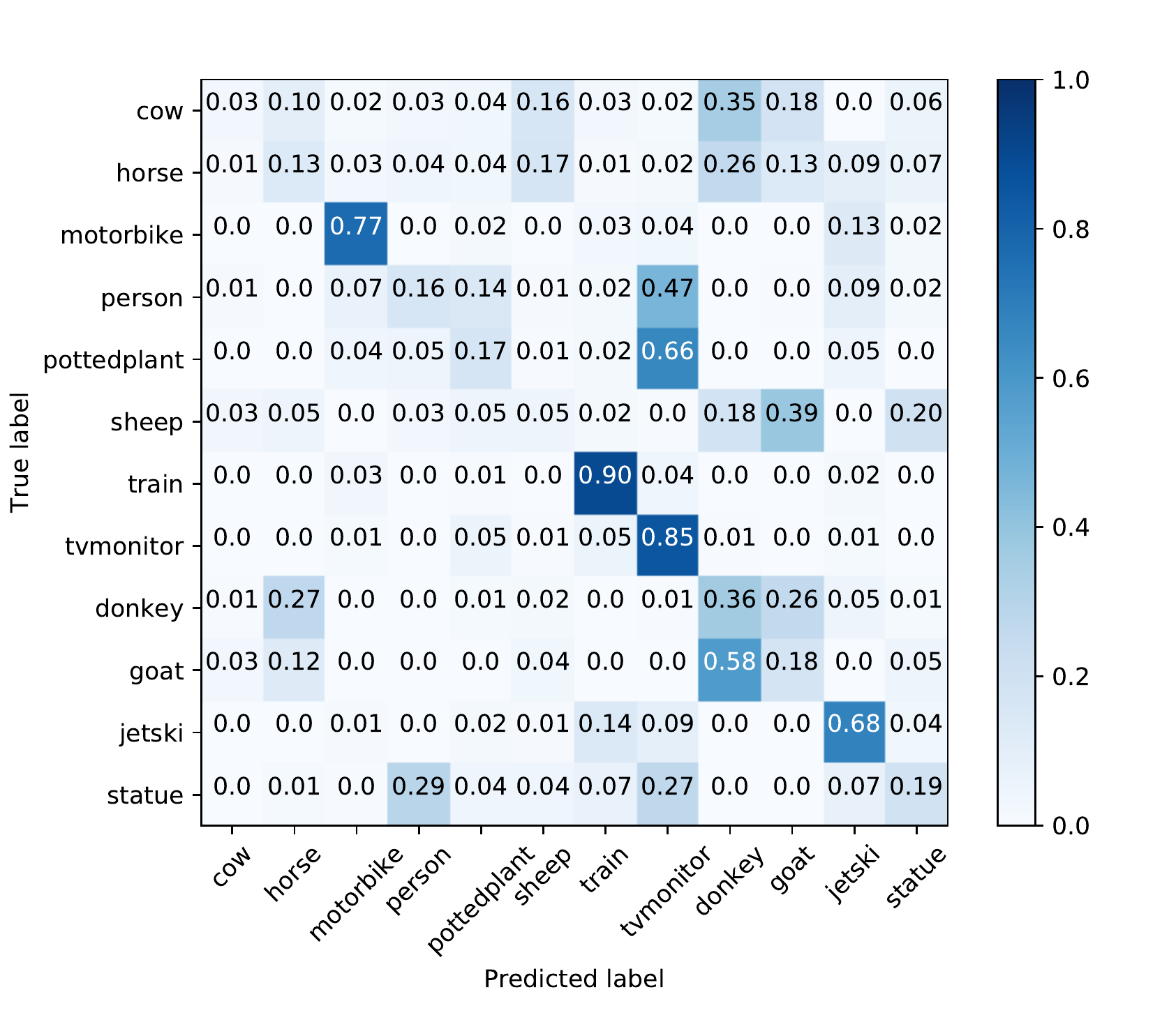}
}
\hspace{-20pt}
\subfigure[Ours]{
\includegraphics[width=0.51\linewidth]{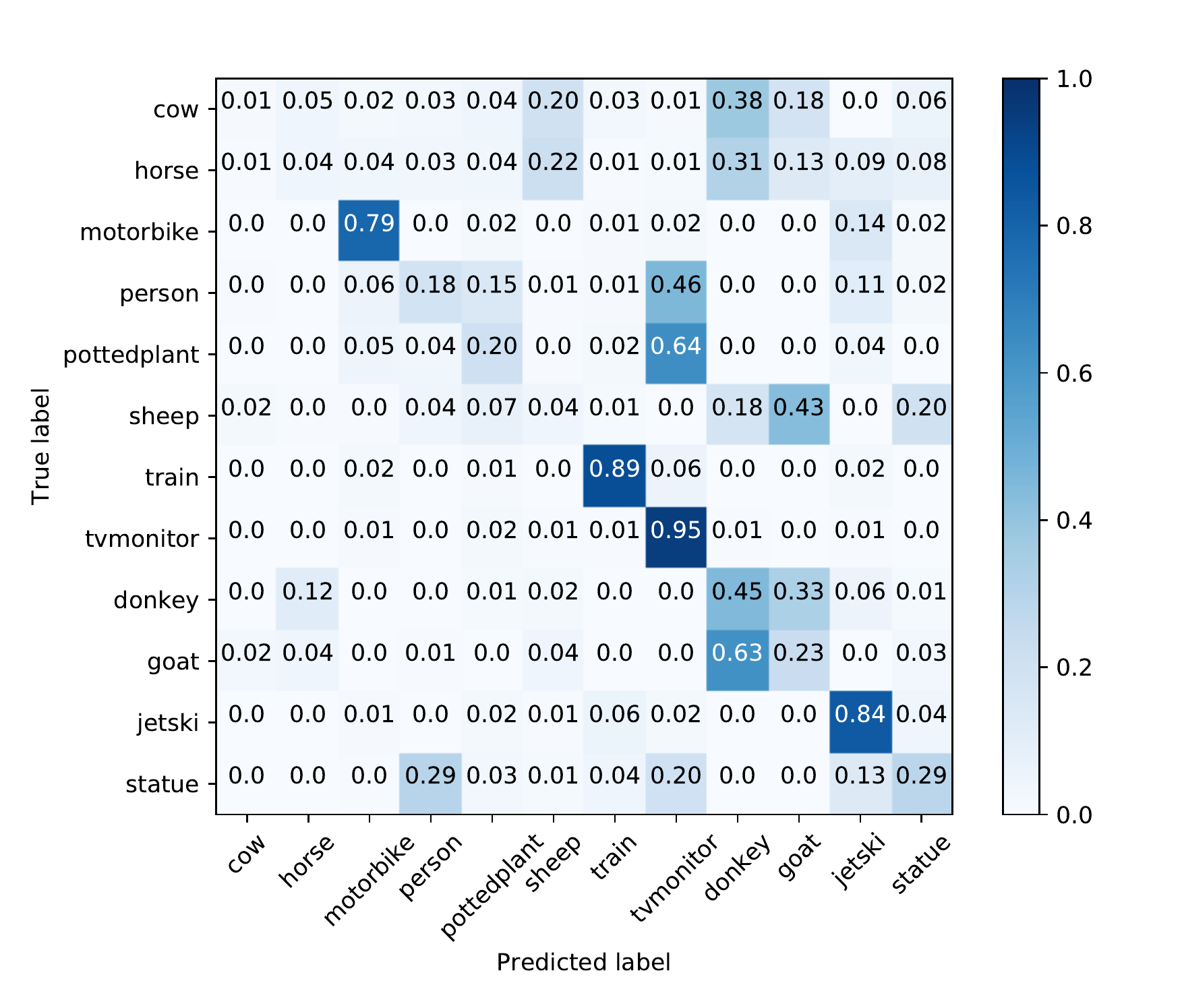}
}
\end{center}
\vspace{-15pt}
\caption{The confusion matrix on the evaluation of aPaY dataset.} 
\label{fig:confusion}
\vspace{-10pt}
\end{figure}

From the results in Table~\ref{tab:gzsl1} and Table~\ref{tab:gzsl2}, we can draw the similar conclusions as from Table~\ref{tab:zsl}. Our approach performs better than existing methods. Our results are significantly better on the unseen samples and harmonic mean, which means our proposed method has a much better generalized ability. It is able to classify the samples into the {\it true} category. Our approach is stably dependable on both seen and unseen classes. Although some previous methods, e.g., DAP, ESZSL and SAE, perform well on the conventional zero-shot learning setting with unseen samples, their performances degrade dramatically on the generalized zero-shot learning. They tend to mess up when the possible categories of unseen samples become large. Thus, the applicability of these methods is limited in real applications.


The harmonic mean is more stable regarding outliers than the arithmetic mean and geometric mean. Thus, from the results reported in Table~\ref{tab:gzsl1} and Table~\ref{tab:gzsl2}, we can also observe that our method is more stable than the compared methods. It avoids extreme results on different evaluations. In terms of the harmonic mean, we achieved up to 2.8\%, 2.7\%, 1.9\%, 2.7\% and 0.8\% improvements on aPaY, AwA, CUB, FLO and SUN, respectively. The average is over the five is 2.2\%. Although our method did not perform the best on some seen categories, it performs almost neck to neck with the previous state-of-the-arts. These results verified the outstanding generalization ability of our method.

Considering the fact that both GAZSL and f-CLSWGAN leverage GANs to synthesize unseen samples, the performance boost of our method can be attributed to two aspects. One is that we introduce soul samples to guarantee that each generated sample is highly related with the semantic description. The soul samples regularizations also address the multi-view characteristic. As a result, it can automatically take care of the domain-shift problem caused by different views in zero-shot learning. The other aspect is that our cascade classifier is able to leverage the results from the first classifier and strengthen the second one. Such a formulation provides the results via a coarse-to-fine manner. The results verify that it is beneficial to leverage the invariant side of generative ZSL. The invariant side regularizations guarantee that each synthesized sample is highly related with the real ones and corresponding semantic descriptions.

\subsection{Model Analysis}
In this section, we analyze our model under different settings. Since our GAN generates {\it visual features} rather than {\it image pixels}, it is inappropriate to show the synthesized results with images. We will analyze our model in terms of the generalization ability and stability. The sensitivity of hyper-parameters are also discussed.

\subsubsection{Class-wise Accuracy}
To show the experimental results of our method in a more fine-grained scale, we report the confusion matrix of f-CLSGAN and our method on the aPaY dataset in Fig.~\ref{fig:confusion}.
Compared with Fig.~\ref{fig:confusion}(a) and Fig.~\ref{fig:confusion}(b), we can see that our method generally has better accuracy on most of the categories. Notably, we can see that the accuracy on category ``tvmonitor'', ``donkey'' and ``jetski'' are boosted around 10\% against f-CLSWGAN. There is also a common phenomenon that the ZSL methods perform poorly on some unseen categories. We will investigate fine-grained / class-wise zero-shot learning in our future work. 


\begin{figure*}[t!h]
\begin{center}
\subfigure[Classification loss ($\lambda$)]{
\includegraphics[width=0.23\linewidth]{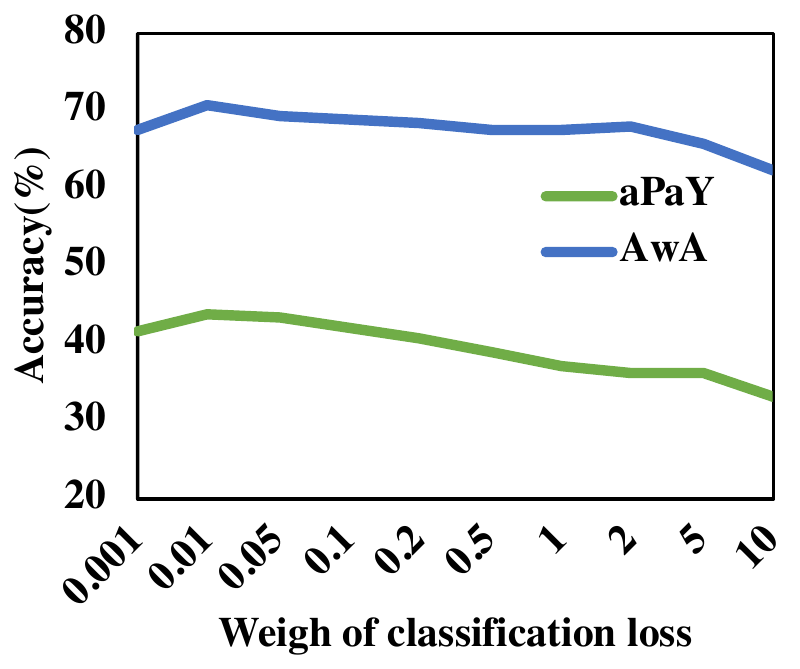}
}
\subfigure[Soul sample regularization]{
\includegraphics[width=0.23\linewidth]{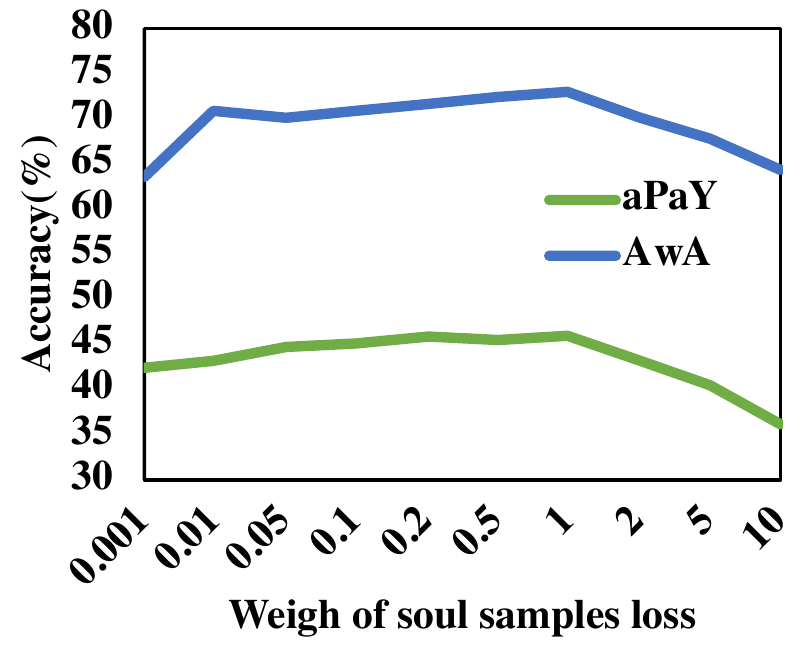}
}
\subfigure[Sample entropy threshold]{
\vspace{-5pt}
\includegraphics[width=0.226\linewidth]{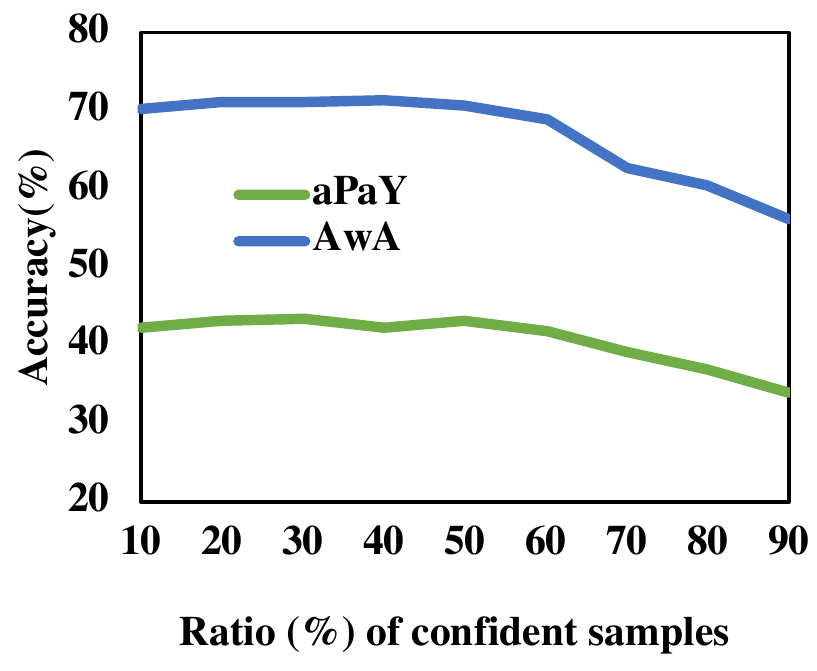}
}
\subfigure[Synthesized sample numbers]{
\includegraphics[width=0.23\linewidth]{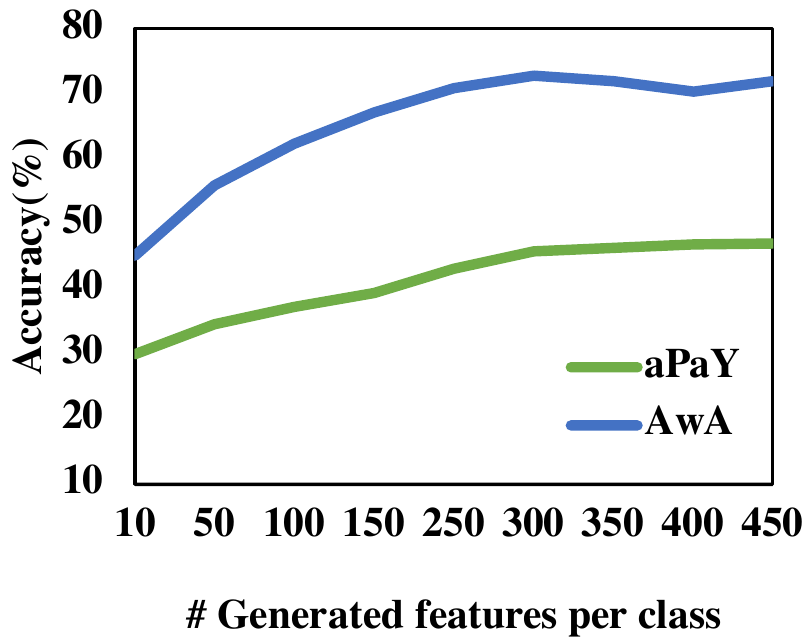}
}
\end{center}
\vspace{-10pt}
\caption{Parameter sensitivity. The horizontal axis of (c) indicates the sample entropy threshold is not larger than the entropy of $x\%$ samples where all sample entropies are sorted from small to large., e.g., $50$ indicates the sample entropy threshold is set as the median of all sample entropies. The horizontal axis of (d) indicates synthesized sample numbers per class.}
\label{fig:para}
\vspace{-10pt}
\end{figure*}

\subsubsection{Parameter Sensitivity}
\begin{figure}[t]
\begin{center}
\includegraphics[width=0.93\linewidth]{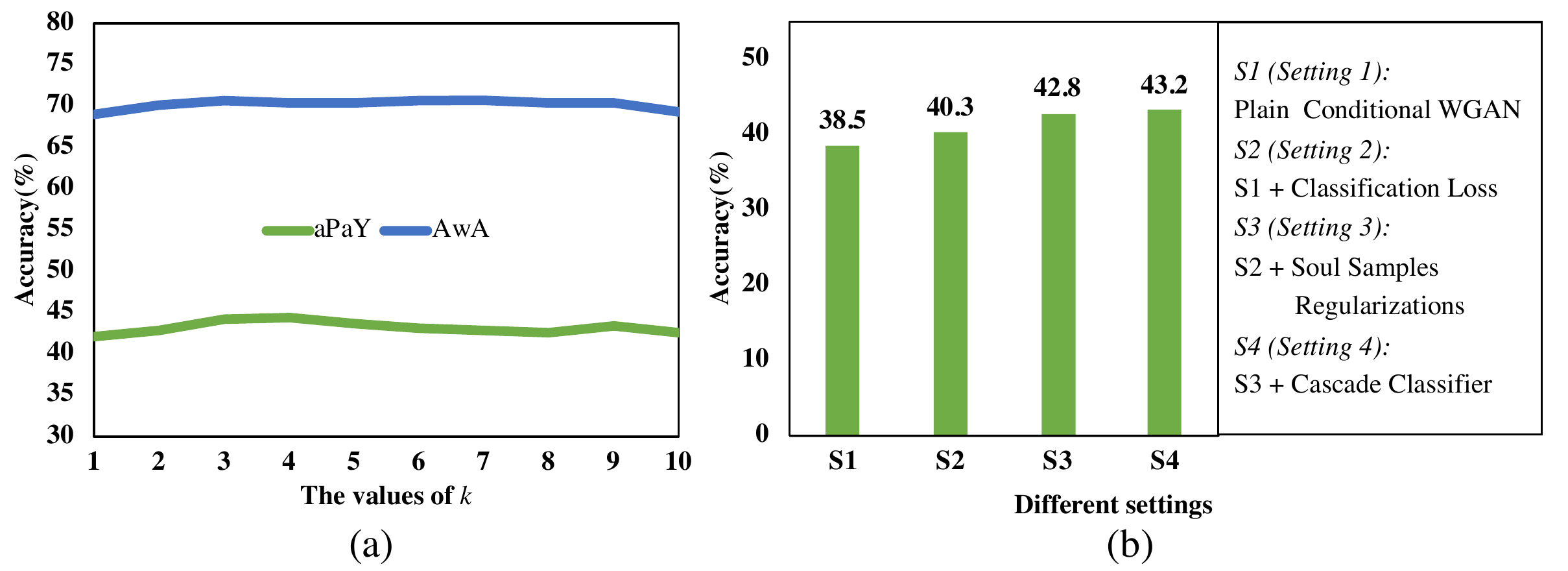}
\end{center}
\vspace{-10pt}
\caption{The results of different $k$ (number of clusters) and ablation analysis of ZSL with aPaY.}
\label{fig:abla}
\vspace{-10pt}
\end{figure}

In our model, we have several hype-parameters to tune. The parameter $\beta$ controls the Lipschitz constraint. As suggested in~\cite{gulrajani2017improved}, we fix $\beta=10$ in this paper. The parameter $\lambda$ balances the supervised classification loss, its influence is reported in Fig.~\ref{fig:para}(a). In our formulation, we also introduced a weight coefficient to adjust the contribution of soul sample regularizations. Its sensitivity is reported in Fig.~\ref{fig:para}(b). Similarly, Fig.~\ref{fig:para}(c) and Fig.~\ref{fig:para}(d) show the effects of sample entropy threshold and synthesized sample numbers per class, respectively. From the results, we can see that the weight parameters for classification loss and soul sample regularization should be relatively small. The sample entropy threshold is recommended to set to be smaller than the median of all samples. The more synthesized samples, the better results generally there will be. However, more samples also introduce more noises and need more training costs. In practice, we suggest to split the seen categories as training set and validation set for cross-validation. Specifically. we report the sensitivity of $k$ in Fig.~\ref{fig:abla}(a). Since $k$ is not sensitive, we fix $k=3$ to reduce the computation cost.

\begin{figure}[t]
\begin{center}
\subfigure[Zero-shot learning]{
\includegraphics[width=0.47\linewidth]{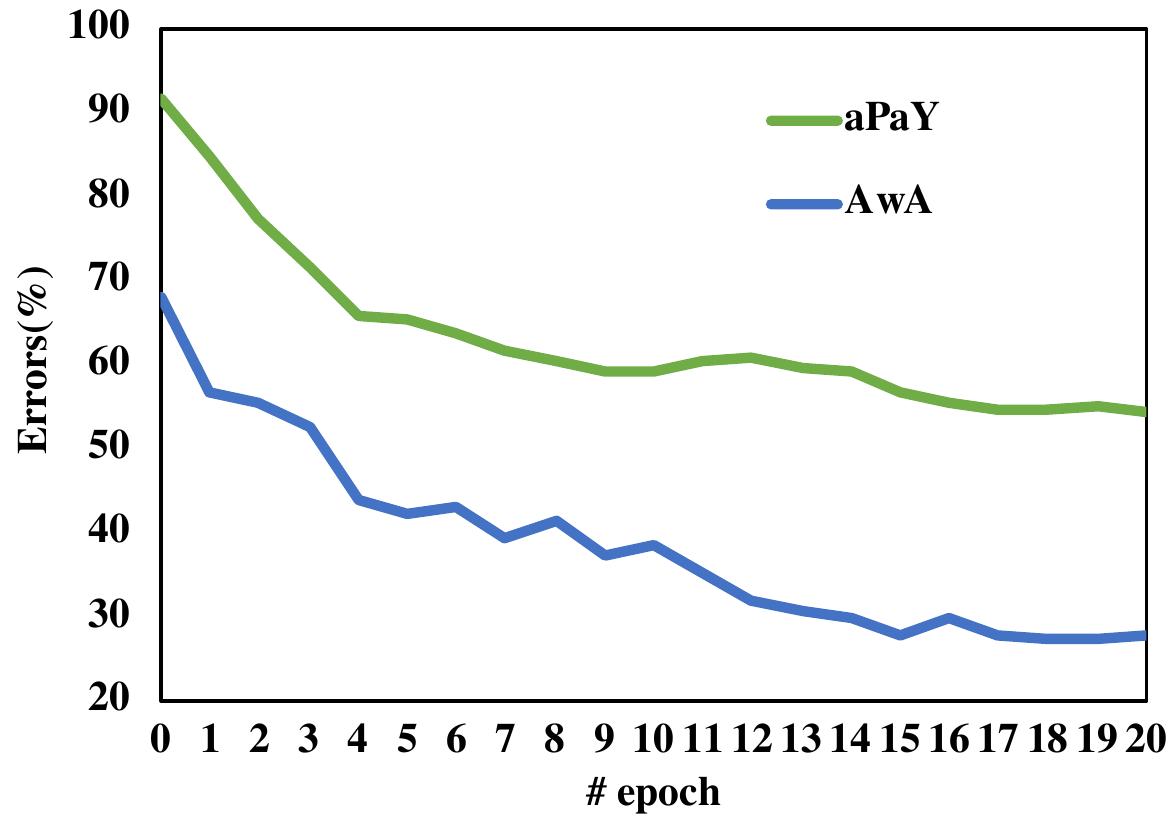}
}
\subfigure[Generalized ZSL]{
\includegraphics[width=0.47\linewidth]{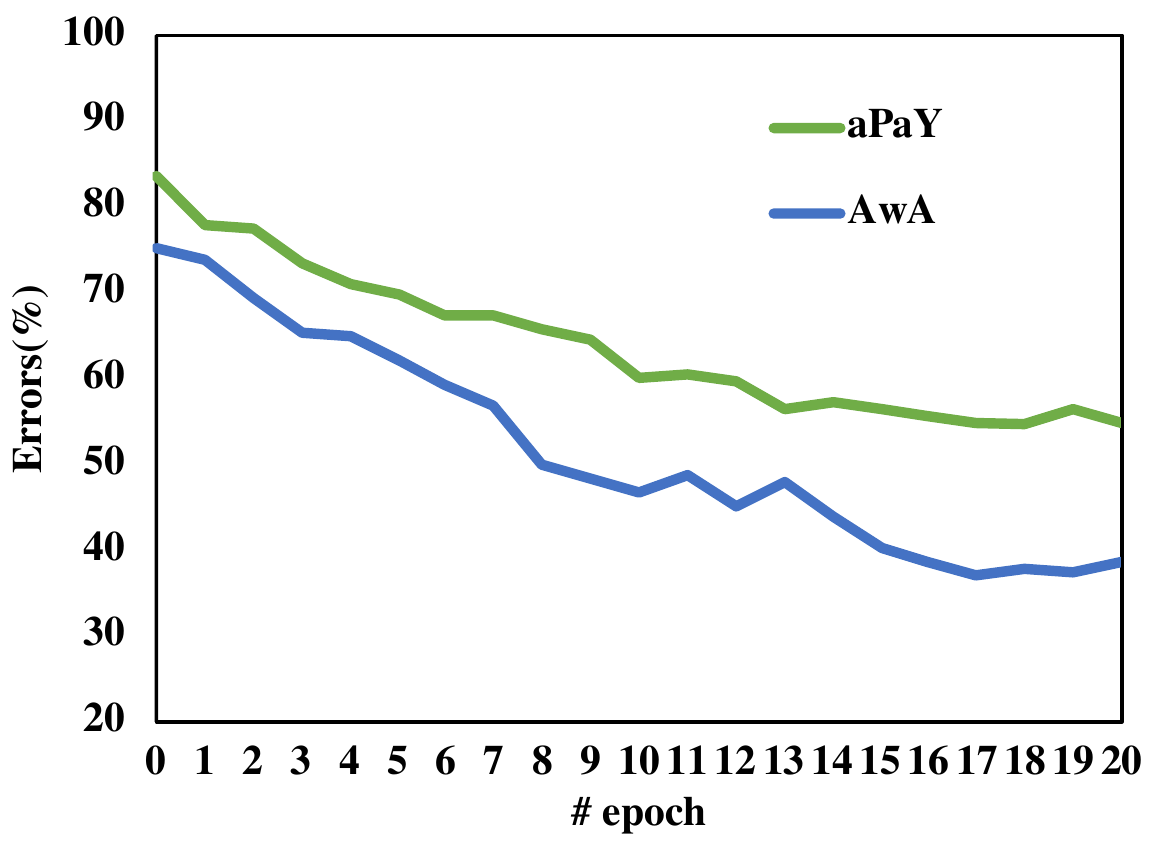}
}
\end{center}
\vspace{-10pt}
\caption{The trends of training stability. For GZSL in (b), we report the harmonic mean on both seen and unseen samples. }
\label{fig:stab}
\vspace{-5pt}
\end{figure}

\subsubsection{Model Stability}
Since our approach deploys an adversarial training manner, it needs several epochs to achieve the balance between the generator and the discriminator. In Fig.~\ref{fig:stab}, we report the zero-shot learning and generalized zero-shot learning results of our method with different epochs in terms of testing error. The results reflect the training stability of our model. It can be seen that our model shows a stable training trend with the increasing of training epochs. Although there are small fluctuations, our model can achieve a stable results with $30$ epochs. For different real-world applications, one can deploy cross-validation to choose the optimal epoch.

\subsubsection{Ablation Analysis}
Conditional WGAN has been a cutting-edge but popular technique in computer vision tasks. It is more like an infrastructure in the community. Thus, we fix the conditional WGAN and focus on soul sample regularization and the cascade classifier in this section. We first report the results of plain conditional WGAN. Then, we introduce additional components into the model and observe the effects of them. The results of ablation analysis are reported in Fig.~\ref{fig:abla}(b). The five settings demonstrate that different components in our framework are all significant. The supervised loss guarantees that the generated features are discriminative. The soul samples regularizations constrain that each synthesized sample is close to the very semantic descriptions. Multiple soul samples per class provide a relaxed solution to handle domain shift problem caused by the multi-view issue. The cascade classifier leverages the result of sample entropy and presents a more fine accuracy.

\section{Conclusion}

In this paper, we propose a novel zero-shot learning method by taking advantage of generative adversarial networks. Specially, we deploy conditional WGAN to synthesize fake unseen samples from random noises. To guarantee that each generated sample is close to real ones and their corresponding semantic descriptions, we introduce soul samples regularizations in the GAN generator. At the zero-shot recognition stage, we further propose to use a cascade classifier to fine-tune the accuracy. Extensive experiments on five popular benchmarks verified that our method can outperform previous state-of-the-art ones with remarkable advances. In our future work, we will explore data augmentation with GAN which can be used to synthesize more semantic descriptions to cover more unseen samples.

\section*{Acknowledgments}
This work was supported in part by the National Natural Science Foundation of China under Grant 61806039, 61832001, 61802236, 61572108 and 61632007, in part by the ARC under Grant FT130101530, in part by the National Postdoctoral Program for Innovative Talents under Grant BX201700045, and in part by the China Postdoctoral Science Foundation under Grant 2017M623006.

{\small
\bibliographystyle{ieee_fullname}
\bibliography{egbib}
}

\end{document}